\documentclass{article} 
\usepackage{iclr2027_conference,times}

\usepackage{amsmath,amsfonts,bm}

\def\eqref#1{equation~\ref{#1}}

\def\1{\bm{1}}

\def\vs{{\bm{s}}}

\DeclareMathAlphabet{\mathsfit}{\encodingdefault}{\sfdefault}{m}{sl}
\SetMathAlphabet{\mathsfit}{bold}{\encodingdefault}{\sfdefault}{bx}{n}

\usepackage{hyperref}
\usepackage{url}

\usepackage{multirow}
\usepackage{booktabs}
\usepackage{subcaption}
\usepackage{graphicx}
\usepackage{xspace}
\usepackage{marvosym}

\newcommand{\method}{\textsc{PhysForge}\xspace}
\newcommand{\videomodule}{\textsc{PhysFit}\xspace}
\newcommand{\compmodule}{\textsc{PhysComp}\xspace}
\newcommand{\videoforce}{Hallucinated Forcing\xspace}

\hypersetup{
  colorlinks=true,
  linkcolor=red!50!black,   
  citecolor=green!50!black, 
  urlcolor=blue!50!black    
}

\title{\method: Material-Controllable 3D Reconstruction with Separable Components}

\author{Yuchen Wang\textsuperscript{\(\dagger\)} \\
{\small School of Computer Science} \\
{\small Nanjing University} \\
{\small Jiangsu, 210023, China} \\
{\small \texttt{602025330026@smail.nju.edu.cn}}
\And
Yanqi Bao\textsuperscript{\(\dagger\ddagger\)} \\
{\small Horizon Robotic} \\
{\small Shanghai, China} \\
{\small \texttt{yanqibao1997@}}\\
{\small \texttt{gmail.com}}
\And
Zheng Gu \\
{\small College of Computer Science} \\
{\small and Software Engineering} \\
{\small Shenzhen University} \\
{\small Guangdong, 518060, China} \\
{\small \texttt{guzheng@szu.edu.cn}}
\And
Zhengqing Chen \\
{\small Horizon Robotic} \\
{\small Shanghai, China} \\
{\small \texttt{zhengqing.chen@horizon.cc}}
\And
Jieqi Shi \\
{\small School of Intelligence Science} \\
{\small and Technology} \\
{\small Nanjing University} \\
{\small Jiangsu, 215163, China} \\
{\small \texttt{isjieqi@nju.edu.cn}}
\And
Jing Huo\textsuperscript{\Letter} \\
{\small School of Computer Science} \\
{\small Nanjing University} \\
{\small Jiangsu, 210023, China} \\
{\small \texttt{huojing@nju.edu.cn}}
}

\begin{document}

\makeatletter
\DeclareRobustCommand\onedot{\futurelet\@let@token\@onedot}
\def\@onedot{\ifx\@let@token.\else.\null\fi\xspace}

\def\eg{\emph{e.g}\onedot} \def\Eg{\emph{E.g}\onedot}
\def\ie{\emph{i.e}\onedot} \def\Ie{\emph{I.e}\onedot}
\def\cf{\emph{c.f}\onedot} \def\Cf{\emph{C.f}\onedot}
\def\etc{\emph{etc}\onedot} \def\vs{\emph{vs}\onedot}
\def\wrt{w.r.t\onedot} \def\dof{d.o.f\onedot}
\def\etal{\emph{et al}\onedot}
\makeatother

\maketitle

\begingroup
\renewcommand{\thefootnote}{\ensuremath{\dagger}}
\footnotetext{Equal contribution}
\renewcommand{\thefootnote}{\ensuremath{\ddagger}}
\footnotetext{Project leader}
\renewcommand{\thefootnote}{\Letter}
\footnotetext{Corresponding author}
\endgroup

\begin{abstract}
Recent advances in physical 3D generation have enabled visually faithful assets with simulatable mechanical behavior. However, existing methods provide limited control over material behavior beyond appearance-based inference and struggle to estimate physical properties from complex interactions between objects and semantic parts. Therefore, we present \method, a Video Diffusion Model (VDM)-based framework for addressing these challenges, comprising two complementary modules for material-controllable physical fitting and component-level interaction modeling. The first, \videomodule, leverages material-conditioned VDM dynamics as references for physical-property fitting, introducing windowed fitting to focus on material-revealing interactions and \videoforce to disentangle nonphysical motion hallucinated by the VDM from material responses. The second, \compmodule decomposes complex interactions into individual objects and semantic components, enabling their physical parameters to be optimized independently and flexibly composed for joint simulation. Experiments on PhysAssets and Google Scanned Objects demonstrate that, compared with existing methods, \method provides stronger control over material-dependent physical responses while supporting component-level physical modeling for complex interactions. It achieves the best CLIP and physical controllability scores on both common and uncommon materials, with qualitative comparisons, component-level ablations, and a user study further supporting these improvements. Our project page is available at \href{https://masterwanna.github.io/PhysForge.html}{https://masterwanna.github.io/PhysForge.html}.
\end{abstract}

\section{Introduction}
\label{sec:introduction}

Recent advances in 3D generation and reconstruction have made it possible to create assets with increasingly realistic geometry and appearance~\citep{xiang2026native,liang2026photo3d}. For physical simulation, visual fidelity alone is insufficient: meaningful interaction also requires modeling how objects made of different materials respond to forces, contact, and deformation. This requires 3D assets to be endowed with physical properties that govern their mechanical behavior and enable physically meaningful responses to interaction~\citep{zhong2024springgaus,jiang2025phystwin}.

Toward this goal, existing methods augment reconstructed or generated 3D assets with physical properties inferred using Vision--Language Models (VLMs) or discriminative models trained on physical-property data~\citep{cao2026physxanythings,cao2026physx3d}. However, two key challenges remain. First, appearance-based inference provides limited control over material behavior. Such predictions are largely shaped by appearance--material correlations learned from training data, making it difficult to assign a user-specified material to the same object and obtain the corresponding physical response~\citep{xu2025gaussianproperty,li2023pacnerf}. Second, existing methods have limited ability to infer physical properties from complex interactions between objects or semantic parts~\citep{lin2025omniphysgs,xu2026neuspring}. In most scenes, material behavior is often manifested through contact and deformation among these interacting entities, whereas existing methods mainly rely on object-level visual observations and struggle to capture such interaction-dependent behavior~\citep{zhang2024physdreamer}.

\begin{figure}[t]
    \centering
    \includegraphics[width=\linewidth]{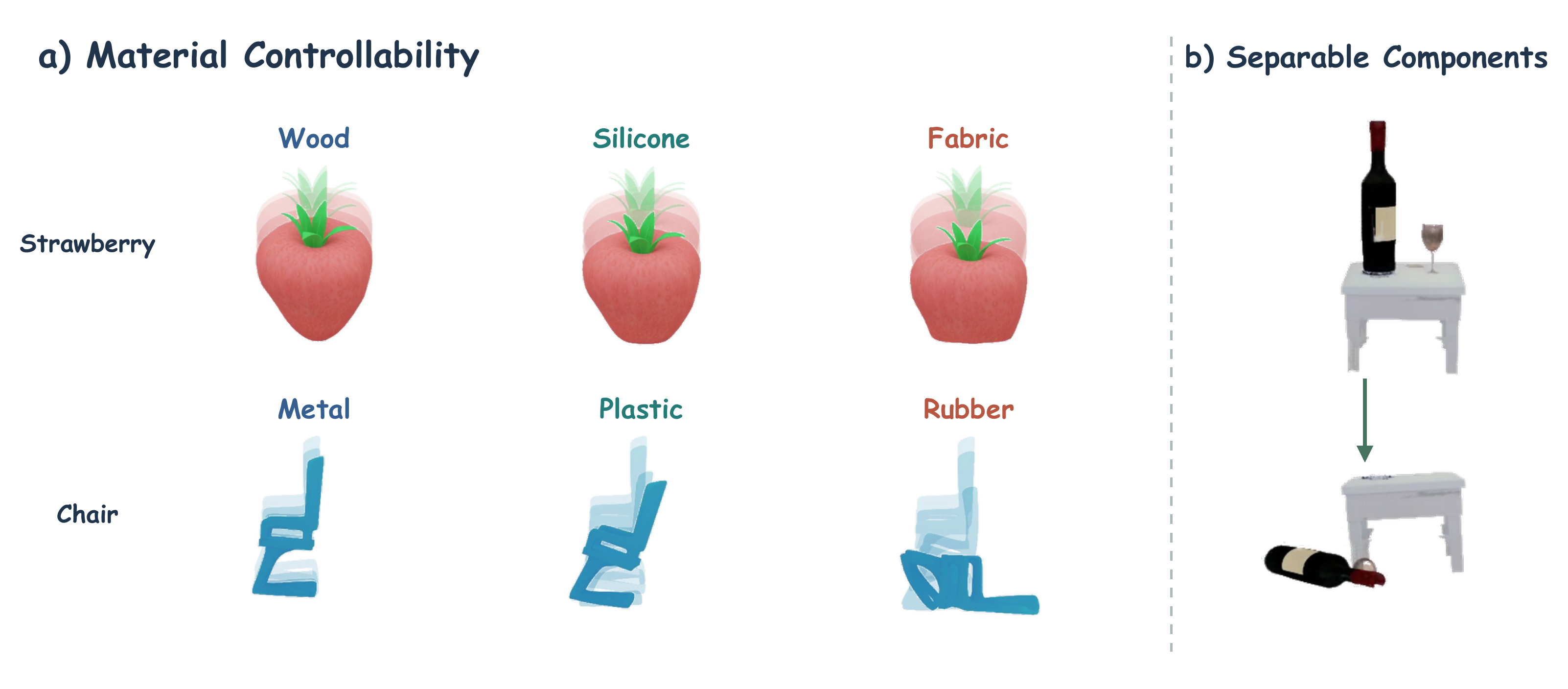}
    \caption{\method achieves material controllability and separable components.}
    \label{fig:teaser}
\end{figure}

To address these limitations, we propose \method, a video diffusion model (VDM)-based framework for constructing physically simulatable 3D assets with separable semantic components, as shown in Figure~\ref{fig:teaser}. Our framework consists of two modules, \videomodule and \compmodule. \videomodule leverages the ability of VDMs to generate material-conditioned dynamics and introduces a robust generated-video fitting mechanism that mitigates the influence of nonphysical motion when estimating physical properties for user-specified materials. \compmodule decomposes complex interactions into individual objects and semantic components, enabling their physical properties to be analyzed separately and flexibly composed for joint simulation.

Specifically, \videomodule exploits the ability of VDMs to generate different material-dependent dynamics for the same object and treats these material-conditioned outputs as reference dynamics for physical-property fitting. To further streamline the fitting process, we develop a windowed fitting paradigm that optimizes physical parameters only over informative intervals that reveal material properties, such as contact, collision, and deformation events. An unavoidable limitation is that VDM-generated dynamics does not fully obey physical laws, making direct fitting to the entire generated sequence unreliable. Therefore, we introduce \textbf{\videoforce} to account for generated motion that cannot be explained solely by the underlying physical parameters while separating it from the material response used for fitting. For complex interactions, \compmodule propagates object--part identities from video to reconstructed 3D geometry, allowing the physical parameters of individual objects and semantic parts to be optimized independently. It then determines simulation groups based on their relative motion, enabling independently optimized components to be composed for joint simulation while preserving their semantic identities.

We conduct extensive experiments on comparing \method to other baseline methods. Qualitative results show that the material controllability and component separability enabled by our design lead to more physically plausible responses and interactions; a human study further finds that participants prefer our results. In quantitative evaluations, \method achieves the best CLIP score and physical controllability score for both common and uncommon materials. Finally, ablation studies demonstrate the effectiveness of \compmodule and \videomodule.

Our contributions are as follows:
\begin{itemize}
    \item We introduce \method, a VDM-based framework for constructing physically simulatable 3D assets with controllable material behavior and separable semantic components, enabling material specification beyond appearance-based physical-property inference.

    \item We propose \videomodule, which leverages material-conditioned VDM dynamics for controllable physical-property estimation and introduces robust generated-video fitting to account for nonphysical motion and focus optimization on physically informative intervals.

    \item We develop \compmodule to enable independent physical-parameter optimization for individual objects and semantic components, followed by flexible composition into simulation groups based on their relative motion.

    \item Extensive experiments show that \method generates visually faithful 3D assets with plausible material-dependent responses. Ablation studies further validate the effectiveness of \videomodule and \compmodule.
\end{itemize}

\section{Related Work}
\label{sec:related_work}

\subsection{Physical-Property Estimation}

\noindent\textbf{Observation-Driven Inference.} In common cases, physical properties can be inferred from an object's appearance or identified from its observed motion and deformation. GaussianProperty uses segmentation and multimodal reasoning to assign physical attributes, while PhysGM jointly predicts geometry and physical parameters~\citep{xu2025gaussianproperty,lv2026physgm}. PhysX-3D and PhysX-Anything further incorporate physical attributes into asset generation and reconstruction~\citep{cao2026physx3d,cao2026physxanythings}. Complementing these visual and semantic cues, PAC-NeRF couples neural reconstruction with continuum simulation for parameter identification, and Spring-Gaus recovers deformable dynamics with a spring-based Gaussian representation~\citep{li2023pacnerf,zhong2024springgaus}. Static images alone provide limited evidence of how an object responds to forces and contact. \method therefore uses material-conditioned motion, collisions, and deformation as dynamic evidence, providing additional constraints for more accurate physical-property estimation.

\noindent\textbf{Video-Prior-Guided Fitting.} Video generation provides dynamic priors for physical modeling. MotionCraft, PhysGen, and PSIVG incorporate simulation guidance~\citep{montanaro2024motioncraft,liu2024physgen,foo2026psivg}, while PhyT2V uses physics-oriented prompt refinement~\citep{xue2025phyt2v}. For physical-property estimation, PhysDreamer fits parameters to generated motion, while DreamPhysics and Physics3D distill video diffusion priors through simulation~\citep{zhang2024physdreamer,huang2025dreamphysics,liu2024physics3d}. PhyMAGIC further incorporates motion-aware frame selection and confidence-guided reasoning~\citep{meng2026phymagic}. Fitting an entire generated sequence can mix material-dependent responses with nonphysical motion, since generated videos can violate physical constraints~\citep{bansal2025videophy}. In particular, we observe \videoforce that produces visually plausible but physically incorrect motion, which can bias the fitted material parameters. \method uses window selection to fit physical-property estimation which limits the influence of this nonphysical motion.

\subsection{Component-Aware 3D Asset Construction}

\noindent\textbf{Semantic decomposition and component reconstruction.} Component-aware reconstruction makes an object's internal structure available for modelling. PartField and P3-SAM recover semantic regions in 3D assets~\citep{liu2025partfield,ma2025p3sam}, while PartCrafter and PartGen generate decomposed geometry with coherent component structure~\citep{lin2026partcrafter,chen2025partgen}. Geometric decomposition alone does not establish how the resulting components respond to physical interactions. \method therefore associates semantic components with their observed dynamic responses, enabling extracted components to retain both their geometry and corresponding physical properties from both semantic recognition and dynamic priors.

\noindent\textbf{Physical modeling of object and component interactions.} PhysGaussian couples Gaussian appearance with continuum mechanics~\citep{xie2024physgaussian}, PhysTwin reconstructs interactive physical twins from observations~\citep{jiang2025phystwin}, and OmniPhysGS learns constitutive behavior under video guidance~\citep{lin2025omniphysgs}. DecoupledGaussian separates objects from contacting scene surfaces to support physical interactions~\citep{wang2025decoupledgaussian}. When multiple objects or components interact, their coupled motion and deformation can obscure which material properties produce the observed response. \method therefore combines separate material estimation with joint physical fitting to account for both individual component behavior and interactions.

\section{Method}
\label{sec:method}

\begin{figure}[!t]
    \centering
    \includegraphics[width=\linewidth]{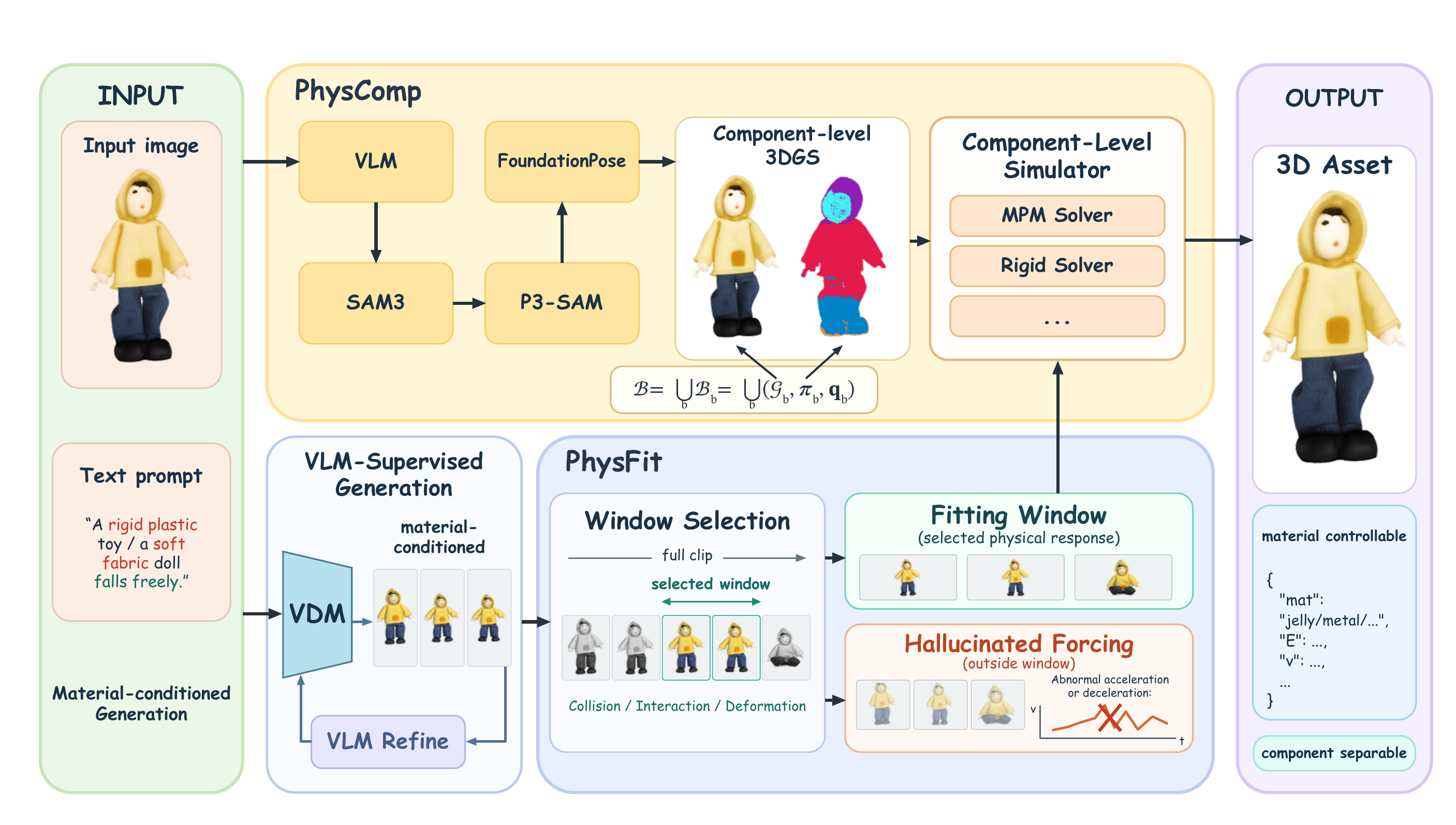}
    \caption{\textbf{Overview of \method.} The framework comprises \videomodule and \compmodule. \videomodule receives a material-conditioned video generated by a VLM-supervised VDM. The generated video is analyzed to select windows for physical-property fitting. \compmodule constructs component-level 3DGS through semantic recognition, 3D reconstruction, component segmentation, and scene placement. Each component is associated with material properties and an initial state, and is then fitted within the windows selected by \videomodule.}
    \label{fig:overview}
\end{figure}

Our goal is to reconstruct physically simulatable 3D assets with controllable material behavior and separable semantic components. To achieve this goal, \method couples two complementary modules: \videomodule uses material-conditioned VDM dynamics to fit physical properties while accounting for nonphysical motion, and \compmodule decomposes the scene into persistent objects and semantic components for separate estimation and joint simulation. Their outputs are combined to produce components that can be composed in a shared physical simulation. We first introduce 3DGS and the particle--grid mechanics underlying the component-level simulator, then present video-guided physical-property fitting and component-aware physical modeling; Figure~\ref{fig:overview} summarizes the framework.

\subsection{Preliminary}
\label{sec:preliminaries}

\subsubsection{3D Gaussian Splatting}

We use 3D Gaussian Splatting (3DGS) as the appearance representation for every reconstructed object~\citep{kerbl2023gaussians}. A 3D Gaussian primitive $g_i=(\boldsymbol\mu_i,\boldsymbol\Sigma_i,\alpha_i,\boldsymbol c_i)$ has center $\boldsymbol\mu_i\in\mathbb{R}^3$, positive-definite covariance $\boldsymbol\Sigma_i$, opacity $\alpha_i$, and view-dependent color coefficients $\boldsymbol c_i$. With rotation $\mathbf R_i$ and diagonal scale $\mathbf S_i$,
\begin{equation}
    \boldsymbol\Sigma_i=\mathbf R_i\mathbf S_i\mathbf S_i^{\top}\mathbf R_i^{\top}.
\end{equation}
For a camera projection $\pi$, the projected covariance is $\boldsymbol\Sigma_i^{2\mathrm D}=\mathbf J_i\boldsymbol\Sigma_i\mathbf J_i^{\top}$, where $\mathbf J_i$ is the projection Jacobian. With $G_i(\mathbf p)$ denoting the projected Gaussian kernel, front-to-back compositing gives
\begin{equation}
    I(\mathbf p)=\sum_i T_i(\mathbf p)\alpha_iG_i(\mathbf p)\mathbf c_i,
    \qquad
    T_i(\mathbf p)=\prod_{j<i}\bigl(1-\alpha_jG_j(\mathbf p)\bigr).
\end{equation}
Each Gaussian stores both an object identifier and a semantic component identifier. A separate mapping assigns these labelled components to simulated bodies, allowing their semantic identities to persist under different simulation groupings.

\subsubsection{Material Point Method}

The material point method (MPM) represents a deformable body by particles that carry position $\mathbf x_p$, velocity $\mathbf v_p$, mass $m_p$, and deformation gradient $\mathbf F_p$~\citep{sulsky1994mpm,hu2018moving}. At each time step, particle-to-grid (P2G) transfer maps particle mass and momentum to grid nodes, where material forces, external forces, and contact update the nodal velocities. Grid-to-particle (G2P) transfer then interpolates the updated grid velocities back to the particles as $\mathbf v_p^{t+1}=\sum_i w_{ip}^t\mathbf v_i^{t+1}$ and advances their positions as $\mathbf x_p^{t+1}=\mathbf x_p^t+\Delta t\,\mathbf v_p^{t+1}$, where $w_{ip}^t$ is the interpolation weight between particle $p$ and grid node $i$, and $\Delta t$ is the time-step duration. The deformation gradient is updated by
\begin{equation}
    \mathbf F_p^{t+1}=\left(\mathbf I+\Delta t\sum_i\mathbf v_i^{t+1}(\nabla w_{ip}^t)^\top\right)\mathbf F_p^t.
\end{equation}
The resulting particle positions, velocities, and deformation gradients become the physical state for the next time step.

\subsection{\videomodule: Video-Guided Physical-Property Fitting}
\label{sec:videomodule}

For a requested material and interaction, a pretrained video diffusion model~\citep{wan2025wan} generates a material-conditioned sequence $\mathcal V=\{I_t\}_{t=0}^{T-1}$. A VLM assesses the generated video and provides feedback to refine the generation~\citep{long2026vista,feng2026newton}. Existing methods fit entire video sequences, including intervals that provide little evidence of the object's physical properties, which leads to suboptimal results. Therefore, \videomodule focuses on selecting physically informative windows and assigning an initial state to each window.

\subsubsection{Windowed Physical-Property Fitting}
\label{sec:window_fitting}

To select physically informative windows, we identify intervals in the generated video where interactions between semantic components produce motion or deformation that reveals their material properties. We treat the scene boundary as a component, so every candidate interaction of $\mathcal C$ is described by the mask intersection between $M_{\mathcal C}$ and the mask $M_{\mathcal C'}$ of another component $\mathcal C'$.

For each candidate pair $(\mathcal C,\mathcal C')$ at frame $t$, we define the contact indicator $\chi_t^{\mathcal C\mathcal C'}=1$ when $M_{t,\mathcal C}\cap M_{t,\mathcal C'}\ne\emptyset$ and zero otherwise. The first transition to $\chi_t^{\mathcal C\mathcal C'}=1$ identifies contact; the window starts at the last frame in the preceding approaching run for which the masks are disjoint and the relative center motion indicates approach, thereby preserving the incoming state. The window ends when the interacting objects come to rest or the contact indicator returns to $\chi_t^{\mathcal C\mathcal C'}=0$.

For such an interaction window $w$, we define its temporal frames as $\mathcal W_w=[t_w^{\mathrm{in}},t_w^{\mathrm{out}}]$. The component set involved in the interaction is $\{\mathcal C\}\cup\bigcup_{t=t_w^{\mathrm{in}}}^{t_w^{\mathrm{out}}}\{\mathcal C'\,|\,\mathcal C'\ne\mathcal C,\ M_{t,\mathcal C}\cap M_{t,\mathcal C'}\ne\emptyset\}$. These components are organized into simulation bodies by \compmodule, and the resulting body set is denoted by $\mathcal B_w$. A complete fitting event can thus be defined as $\bigl(\mathcal W_w,\mathcal B_w\bigr)$, which is used to fit the physical parameters through joint material refinement in Section~\ref{sec:joint_fitting}.

\subsubsection{Reference Motion and \videoforce}

Window selection determines where material parameters are fitted, but does not by itself provide a reference-consistent incoming state. We therefore recover the reference motion explicitly: for each semantic object, SAM3~\citep{carion2026sam3} provides the temporally aligned mask and Depth Anything~\citep{lin2025depthanything3} provides the corresponding depth, from which we track the center trajectory $\widehat{\mathbf c}_{t}$. Given the video frame rate $\mathrm{fps}$, the frame interval is $\Delta t=1/\mathrm{fps}$, the tracked velocity is $\widehat{\mathbf v}_{t}=(\widehat{\mathbf c}_{t+1}-\widehat{\mathbf c}_{t})/\Delta t$, and the apparent acceleration used to reproduce the next center is $\mathbf a^{\mathrm{ref}}_{t}=(\widehat{\mathbf c}_{t+1}-\mathbf c_{t}-\Delta t\mathbf v_{t})/(\Delta t)^2$.

Based on the above analysis, we use \videoforce ($\mathbf a^{\mathrm{ref}}_{t}$) to denote abnormal velocity changes introduced by a nonphysical generation process, such as unexplained acceleration or deceleration. Applying the corresponding velocity update produces a reference trajectory whose tracked positions and velocities follow the generated video, while the nonphysical motion remains distinguishable from the material response used for fitting.

\subsection{\compmodule: Component-Aware Physical Modeling}
\label{sec:compmodule}

Existing methods often simulate each segmented object as a single body, suppressing relative motion and interactions among its semantic components. Therefore, \compmodule focuses on component-level identification and segmentation, organizing the resulting components into simulation bodies to preserve their distinct motion and physical interactions.

\subsubsection{Object--Component Tracking, Reconstruction, and Placement}

To obtain component-level semantic information, a vision--language agent converts generated frames into object-level and component inventories. SAM3~\citep{carion2026sam3} initializes and propagates the corresponding masks across frames and views, preserving identities for repeated or attached components. The masked views are converted into a 3DGS representation $\mathcal G$ using SAM3D~\citep{chen2026sam3d}, TRELLIS~\citep{xiang2025trellis}, or other 3D generation or reconstruction backends. Depth Anything~\citep{lin2025depthanything3} provides depth and camera-consistent scale; P3-SAM~\citep{ma2025p3sam} decomposes the reconstructed geometry into semantic component regions and transfers their labels to the corresponding Gaussians. FoundationPose~\citep{wen2024foundationpose} restores the object pose and scale and replaces the labelled representation in the shared scene coordinate system, yielding a 3DGS asset from which each semantic component can be extracted.

\subsubsection{Simulation Body Assignment and Joint Material Refinement}
\label{sec:joint_fitting}

To prepare the reconstructed semantic components for physical simulation, we organize them into simulation bodies according to their attachment relationships and relative motion. Each simulation body is represented as $\mathcal B_b=(\mathcal G_b,\boldsymbol\pi_b,\mathbf q_b)$, where $\mathcal G_b$ is its labelled 3DGS subset, $\boldsymbol\pi_b$ is its material--physical-parameter pair, and $\mathbf q_b$ is its dynamic state. The labelled Gaussians are converted to surface and interior particles, all bodies are assembled in the shared component-level simulator, and each window initializes $\mathbf q_b$ before the state is propagated through the response.

To fit the observed physical event within each window and discourage interpenetration, we use a physical regulation loss. Let $c_{w,b}$ and $\widehat c_{w,b}$ denote the observed and simulated deformation responses, and let $T_w$, $\widehat T_w$, and $p_w$ denote the observed contact duration, simulated contact duration, and simulated penetration. The physical regulation loss is
\begin{equation}
    \mathcal L_{\mathrm{phys}}=\sum_{w,b}(\widehat c_{w,b}-c_{w,b})^2
    +\lambda_T\sum_w(\widehat T_w-T_w)^2
    +\lambda_P\sum_wp_w^2.
    \label{eq:physical_regulation}
\end{equation}
We further use the initial material estimate predicted by VLM as a prior while refining physical parameters from collision responses in the video. Let $\boldsymbol\theta_b$ denote the physical-parameter vector in $\boldsymbol\pi_b$ and $\boldsymbol\theta_b^0$ its initial estimate. The parameter regularizer softly penalizes deviations from this prior:
\begin{equation}
    \mathcal L_{\mathrm{reg}}=\sum_b\left\|\boldsymbol\theta_b-\boldsymbol\theta_b^0\right\|_2^2.
    \label{eq:parameter_regularization}
\end{equation}
A low weight on this term discourages unreasonable parameter changes caused by noise or insufficient observations while allowing deviations from the initial estimate when supported by the observed response. We optimize the weighted total loss
\begin{equation}
    \mathcal L_{\mathrm{total}}=\lambda_{\mathrm{phys}}\mathcal L_{\mathrm{phys}}+\lambda_{\mathrm{reg}}\mathcal L_{\mathrm{reg}}.
    \label{eq:material_objective}
\end{equation}

\section{Experiments}
\label{sec:experiments}

\subsection{Experimental Setup}
\label{sec:experimental_setup}
\noindent\textbf{Datasets and Implementation.} We evaluate \method on PhysAssets~\citep{lv2026physgm} and Google Scanned Objects (GSO)~\citep{downs2022googlescannedobjects}. PhysAssets contains over 24,000 3D assets annotated with physical properties and accompanied by guiding videos, providing diverse objects for evaluating material-dependent dynamics. GSO provides over 1,000 high-quality 3D scans of household objects, capturing realistic geometry and appearance across diverse object categories. Our framework is implemented on two NVIDIA RTX 4090 GPUs.

\begin{table}[t]
\centering
\caption{Comparison between \method and baseline methods.}
\label{tab:quantitative_comparison}
\small
\setlength{\tabcolsep}{4pt}
\begin{tabular}{lcccc}
\toprule
\multirow{2}{*}{Method} & \multicolumn{2}{c}{Common Materials} & \multicolumn{2}{c}{Uncommon Materials} \\
\cmidrule(lr){2-3}\cmidrule(lr){4-5}
& CLIP $\uparrow$ & Physical Controllability $\uparrow$ & CLIP $\uparrow$ & Physical Controllability $\uparrow$ \\
\midrule
DreamPhysics & 0.250 & 3.2 & 0.248 & 2.6 \\
PSIVG & 0.233 & 2.0 & 0.241 & 1.8 \\
PhysRVG & 0.243 & 3.8 & 0.232 & 3.0 \\
WMReward & 0.246 & 4.4 & 0.244 & 2.6 \\
DeforM & 0.242 & 4.4 & 0.235 & 3.2 \\
PhysGM & 0.246 & 3.8 & 0.247 & 1.8 \\
\midrule
\method & \textbf{0.258} & \textbf{4.6} & \textbf{0.264} & \textbf{4.0} \\
\bottomrule
\end{tabular}%
\end{table}

\begin{figure}[t]
    \centering
    \includegraphics[width=\linewidth]{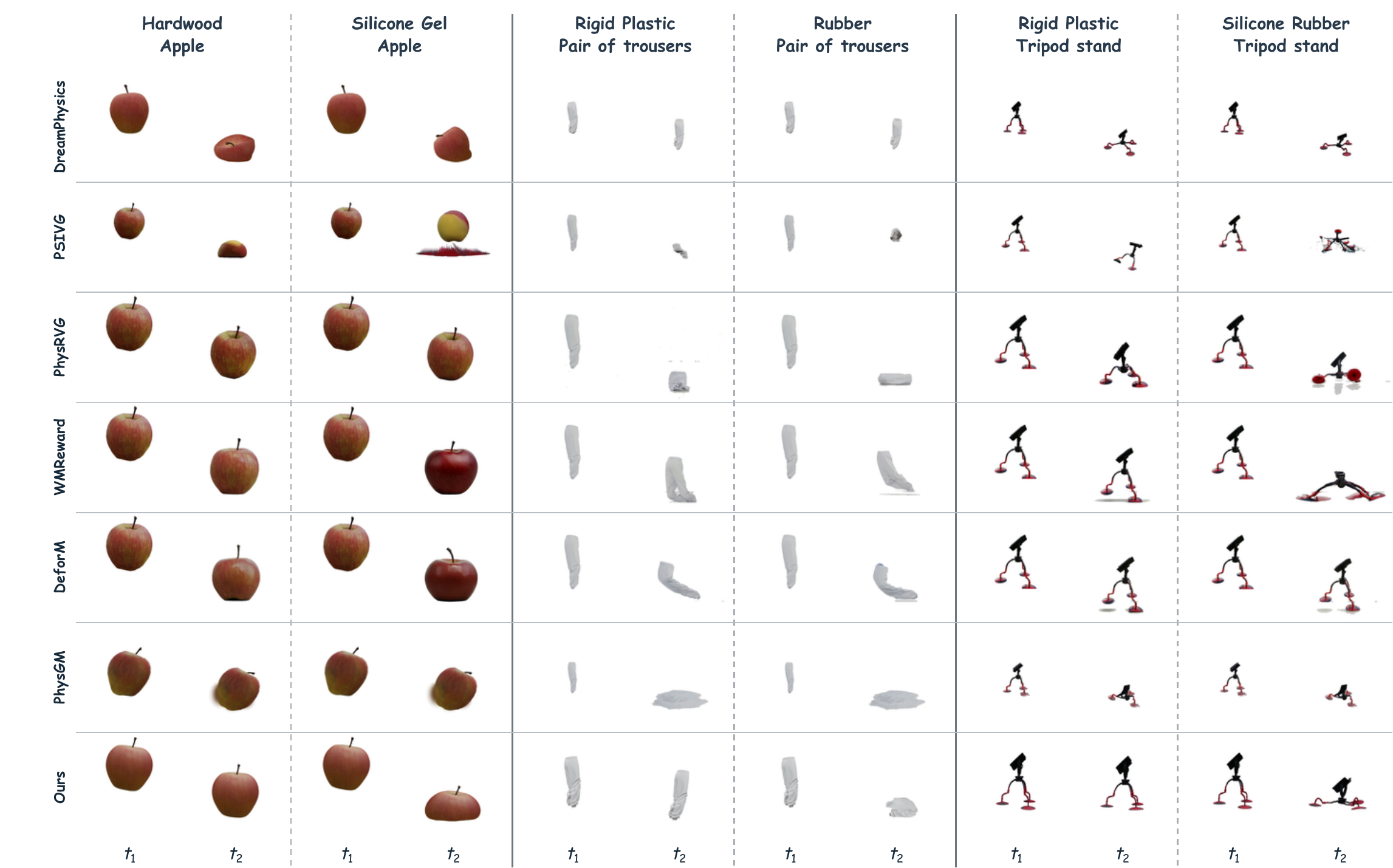}
    \vspace{-5mm}
    \caption{Comparison between \method and baseline methods on material conditions.}
    \vspace{-5mm}
    \label{fig:qualitative_comparison}
\end{figure}

\noindent\textbf{Baselines.} Our qualitative and quantitative comparisons include 6 baselines: DreamPhysics~\citep{huang2025dreamphysics}, PSIVG~\citep{foo2026psivg}, PhysRVG~\citep{zhang2026physrvg}, WMReward~\citep{yuan2026wmreward}, DeforM~\citep{li2026deform}, and PhysGM~\citep{lv2026physgm}. DreamPhysics learns material properties from video diffusion priors through Score Distillation Sampling (SDS) and MPM simulation. PhysGM jointly predicts 3D Gaussians and physical properties from multiview images, whereas PSIVG guides video generation with simulated trajectories reconstructed from an initial generated video. DeforM uses VLM reasoning and spatial--temporal masks to focus generation on regions undergoing physical deformation. PhysRVG incorporates physics-aware reinforcement learning to improve collision dynamics, and WMReward uses a latent world model to score and steer candidate denoising trajectories at inference time.

\begin{figure}[t]
    \centering
    \includegraphics[width=\linewidth]{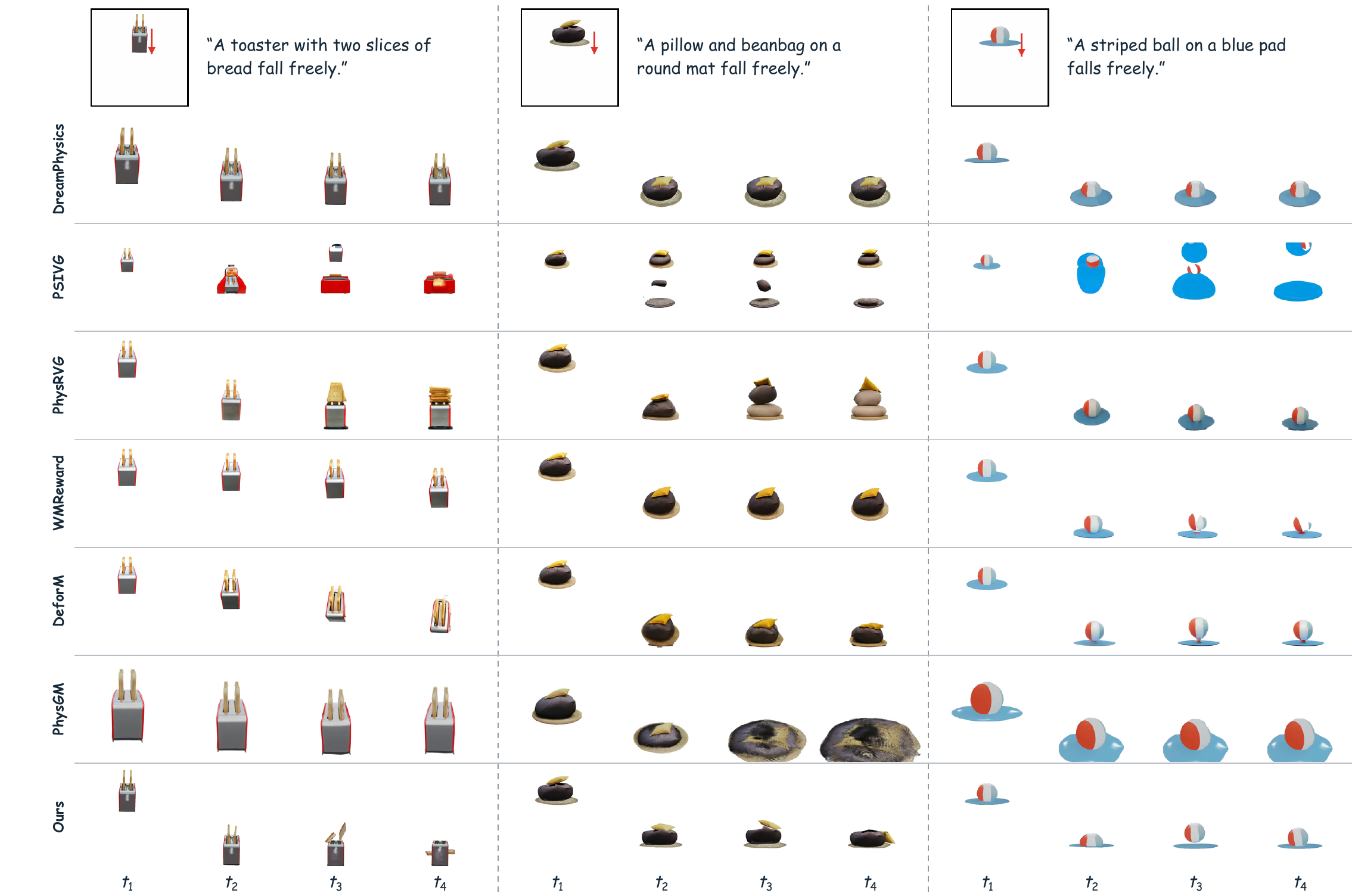}
    \vspace{-5mm}
    \caption{Comparison between \method and baseline methods on component separation.}
    \vspace{-5mm}
    \label{fig:qualitative_separation}
\end{figure}

\noindent\textbf{Simulation.} Our component-level simulator uses Taichi-based MPM and rigid-body solvers~\citep{hu2018moving,hu2019taichi}. For rigid bodies, contact friction controls sliding and restitution controls rebound while the shape remains fixed. For MPM, we generate 150-frame videos at 30 FPS with a maximum of 8 substeps per frame.  Elastic bodies use Young's modulus to control stiffness and Poisson's ratio to control lateral deformation under compression or stretching. Sand uses Young's modulus and Poisson's ratio for its elastic response and an internal friction angle for Drucker--Prager yielding, which determines its resistance to shearing and spreading. Liquids use a weakly compressible fluid model in which the bulk modulus controls the pressure response to compression.

\noindent\textbf{Evaluation Metrics.} We evaluate text alignment, physical material controllability, and human preference. The CLIP score uses ViT-L/14 to encode the text prompt and video frames, and measures their cosine similarity averaged over sampled frames to assess semantic agreement with the requested content. Physical controllability (PC), rated on a five-point scale by VLM, assesses whether the simulated motion, contact, and deformation express the requested material behavior. For the user study, each trial presents seven anonymized videos from \method, DreamPhysics~\citep{huang2025dreamphysics}, PSIVG~\citep{foo2026psivg}, PhysRVG~\citep{zhang2026physrvg}, WMReward~\citep{yuan2026wmreward}, DeforM~\citep{li2026deform}, and PhysGM~\citep{lv2026physgm} in randomized order, together with the text prompts and input images. We also conduct user study on qualitive results. Participants rank the videos by how well they match the requested material while maintaining plausible dynamics.

\subsection{Results and Analysis}
\label{sec:results_analysis}
\noindent\textbf{Qualitative Comparison with Baselines.}\label{sec:qualitative_evaluation} We compare \method with the baselines on material conditions in Figure~\ref{fig:qualitative_comparison} and 
component separation in Figure~\ref{fig:qualitative_separation}. Figure~\ref{fig:qualitative_comparison} shows that the baselines often exhibit similar physical responses under different material conditions. In the tripod case, although PhysRVG exhibits different material behaviors, it fails to maintain appearance consistency. In the silicone-gel apple case, WMReward and DeforM show some deformation, but the apple's appearance also changes. By fitting physical parameters within selected windows, \videomodule extracts material-conditioned physical priors from VDM-generated videos while excluding nonphysical motion, yielding 3D assets whose simulated behavior better matches the requested material.

Figure~\ref{fig:qualitative_separation} shows that the baselines struggle to preserve the separability and relative motion of components, often treating them as a single body during motion and losing their relative movement. PhysGM, which infers material from the input image alone, incorrectly assigns sand-like behavior to the pillow. PhysRVG shows some relative motion in the same case, but introduces an extra component. PSIVG generates multiple apparent components but fails to preserve their original appearance, making its output inconsistent with the input image and prompt. \compmodule identifies objects and segments their semantic components, allowing each component to retain its own physical properties and motion state during simulation which produces more realistic interactions.

\noindent\textbf{Quantitative Comparison with Baselines.}\label{sec:quantitative_evaluation} Table~\ref{tab:quantitative_comparison} reports CLIP scores and PC scores for common and uncommon materials. \method leads all four reported metrics, with an average PC of 4.3 versus 3.8 for the strongest baseline and an average CLIP score of 0.261 versus 0.249 for the strongest baseline. Baseline methods such as WMReward and DeforM show high physical controllability for common materials, however, their performance drops sharply for uncommon user-specified materials. The user study in Table~\ref{tab:user_study} also reflects the advantages of \method in material controllability and component separability. \videomodule selects physically informative fitting windows, while \compmodule segments component-level 3D assets; together, they enable \method to produce physically controllable 3D assets with separable components.

\noindent\textbf{Ablation Studies.}\label{sec:ablation_studies} We conduct two ablations on our key mechanisms: \videomodule and \compmodule.

\begin{table}[t]
    \centering
    \caption{User-study rankings across \method and baseline methods.}
    \label{tab:user_study}
    \small
    \setlength{\tabcolsep}{3pt}
    \begin{tabular}{lccccccc}
        \toprule
         & DreamPhysics & PSIVG & PhysRVG & WMReward & DeforM & PhysGM & \method \\
        \midrule
        Win Rate (\%) $\uparrow$ & 5.56 & 0.56 & 8.33 & 12.22 & 2.78 & 10.00 & \textbf{60.56} \\
        Average Rank $\downarrow$ & 4.2778 & 6.2167 & 3.9500 & 3.5444 & 3.6833 & 4.4222 & \textbf{1.9056} \\
        \bottomrule
    \end{tabular}%
\end{table}

\begin{figure}[t]
    \centering
    \begin{subfigure}[t]{0.49\linewidth}
        \centering
        \includegraphics[width=\linewidth]{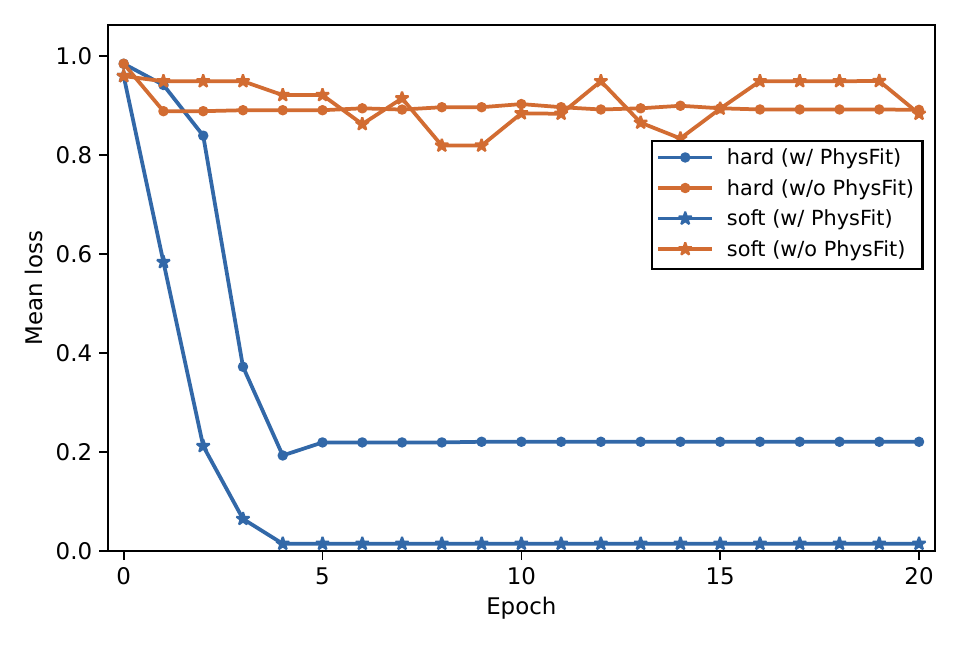}
        \caption{Ablation study for \videomodule.}
        \label{fig:ablation_window_loss}
    \end{subfigure}\hfill
    \begin{subfigure}[t]{0.49\linewidth}
        \centering
        \includegraphics[width=\linewidth]{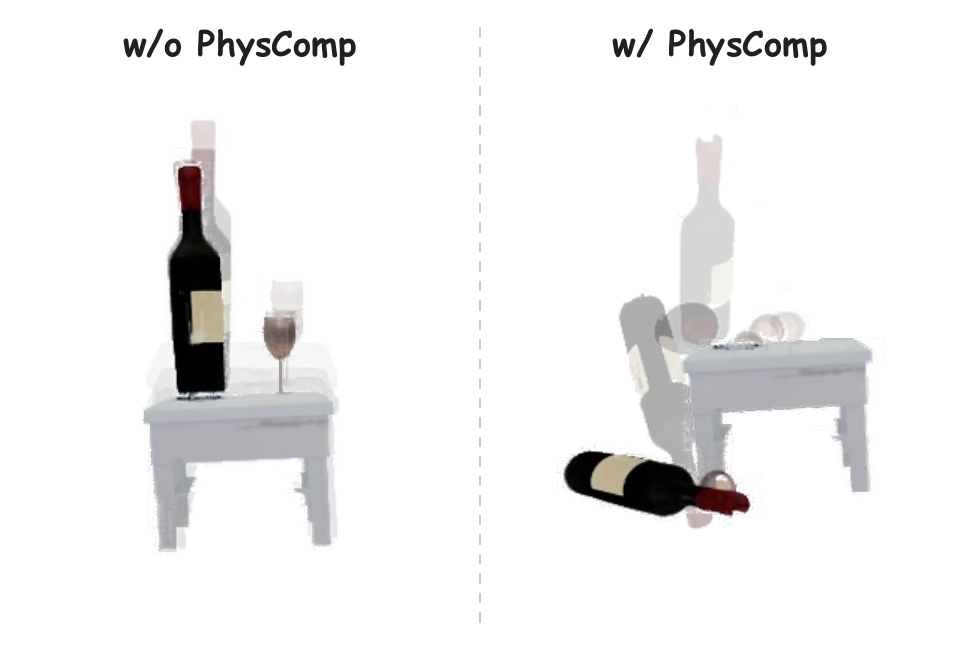}
        \caption{Ablation study for \compmodule.}
        \label{fig:ablation_parts}
    \end{subfigure}
    \caption{Ablation study for \method.}
    \label{fig:ablation_comparison}
\end{figure}


\noindent \raisebox{0.15em}{\rule{0.2em}{0.2em}} \videomodule.\label{sec:ablation_forcing} We remove \videomodule, fitting physical parameters to the entire generated video. We compare quantitative fitting error and convergence speed using loss-versus-optimization-step curves in Figure~\ref{fig:ablation_window_loss}. Removing \videomodule prevents the fitting process from focusing on the windows that reveal physical properties. Over the longer sequence, the loss function struggles to identify the intervals that need to be fitted, which reduces fitting performance and increases fitting time. Therefore, \videomodule enables better physical-property fitting for 3D assets.

\noindent \raisebox{0.15em}{\rule{0.2em}{0.2em}} \compmodule.\label{sec:ablation_components} Removing \compmodule merges semantically distinct components into a single simulated body, as shown in Figure~\ref{fig:ablation_parts}. Viewed as one connected object, the relative motion and interactions among components disappear, making the dynamics physically implausible. With \compmodule, table, bottle and glass retain their own physical properties and motion state. These components that appear connected yet have distinct semantic identities can then collide and wobble upon contact, revealing their individual physical behavior and yielding more plausible interactions.

\section{Conclusion}

In this paper, we analyzed existing methods and found that insufficient material controllability and semantic component separability in reconstructed 3D assets lead to unrealistic physical simulations. To address these limitations, we proposed \method with two key modules: \videomodule and \compmodule. \videomodule improved the efficiency of physical-property fitting through response-window selection and handling of \videoforce, while \compmodule used dynamic and semantic information to decompose reconstructed 3D assets so that individual components retain distinct material properties. Together, these modules enable the reconstruction of different material-conditioned 3D assets from the same input object under different material prompts, while keeping their semantic components extractable. Quantitative and qualitative evaluations show that material controllability and component separability yield more realistic simulations of 3D assets than existing methods. Ablation studies further confirm the effectiveness of both modules.

However, \method still has some limitations. \compmodule uses generated videos to segment objects based on their dynamic and semantic information. Changing the material condition requires additional video generation and physical fitting for each material-condition change. The performance of \method also depends on how well the generated video represents physical behavior, although \videomodule can mitigate hallucinations from video generation to some extent. In future works, more advanced non-physical motion separation techniques can be employed to provide a cleaner window with less nonphysical motion for physical parameter fitting.

\bibliography{iclr2027_conference}
\bibliographystyle{iclr2027_conference}

\appendix
\section{Appendix}
\subsection{Implementation Details}

We use Qwen3.5-4B for object and component analysis and initial material estimation, with SAM3 providing framewise masks. Depth Anything 3 processes at most 24 images per inference call, including shared reference frames spanning the video. Each chunk is aligned to these references using high-confidence depth pixels, and overlapping predictions are blended. Physical simulation of MPM uses a shared grid with 100--300 cells per axis, selected according to scene extent, object size, particle spacing, and motion constraints. Each video frame is integrated with eight substeps, giving $\Delta t=1/(8f)$ for frame rate $f$. Material fitting runs for up to 20 rounds over the detected collision events, initialized from the positions and velocities obtained by video-force fitting. Young's modulus is updated from deformation errors, while candidate selection also accounts for contact duration, penetration, and a material-prior regularizer weighted by $0.01$. Particle volumes are assigned by dividing each body's particle-cloud convex-hull volume by its particle count, and particle masses follow from the assigned density. Ground contact uses a planar boundary with restitution $0.5$ and Coulomb friction coefficient $0.2$.

\subsection{More Results on Material-Controllable}

\begin{figure}[htbp]
    \centering
    \includegraphics[width=\linewidth,keepaspectratio]{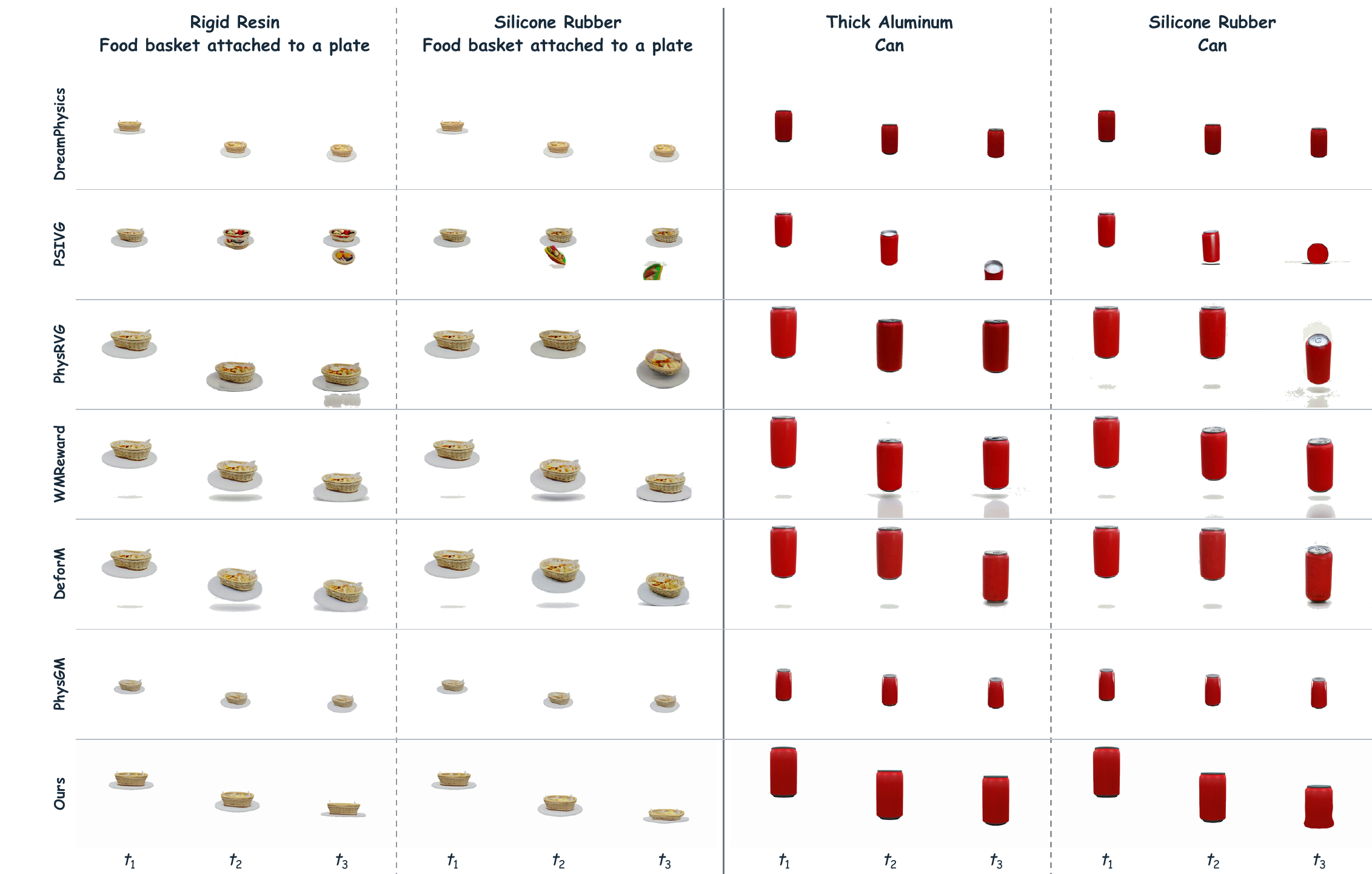}
    \caption{Additional material-controllable comparisons.}
    \label{fig:appendix_controllable_01}
\end{figure}

\begin{figure}[htbp]
    \centering
    \includegraphics[width=\linewidth,keepaspectratio]{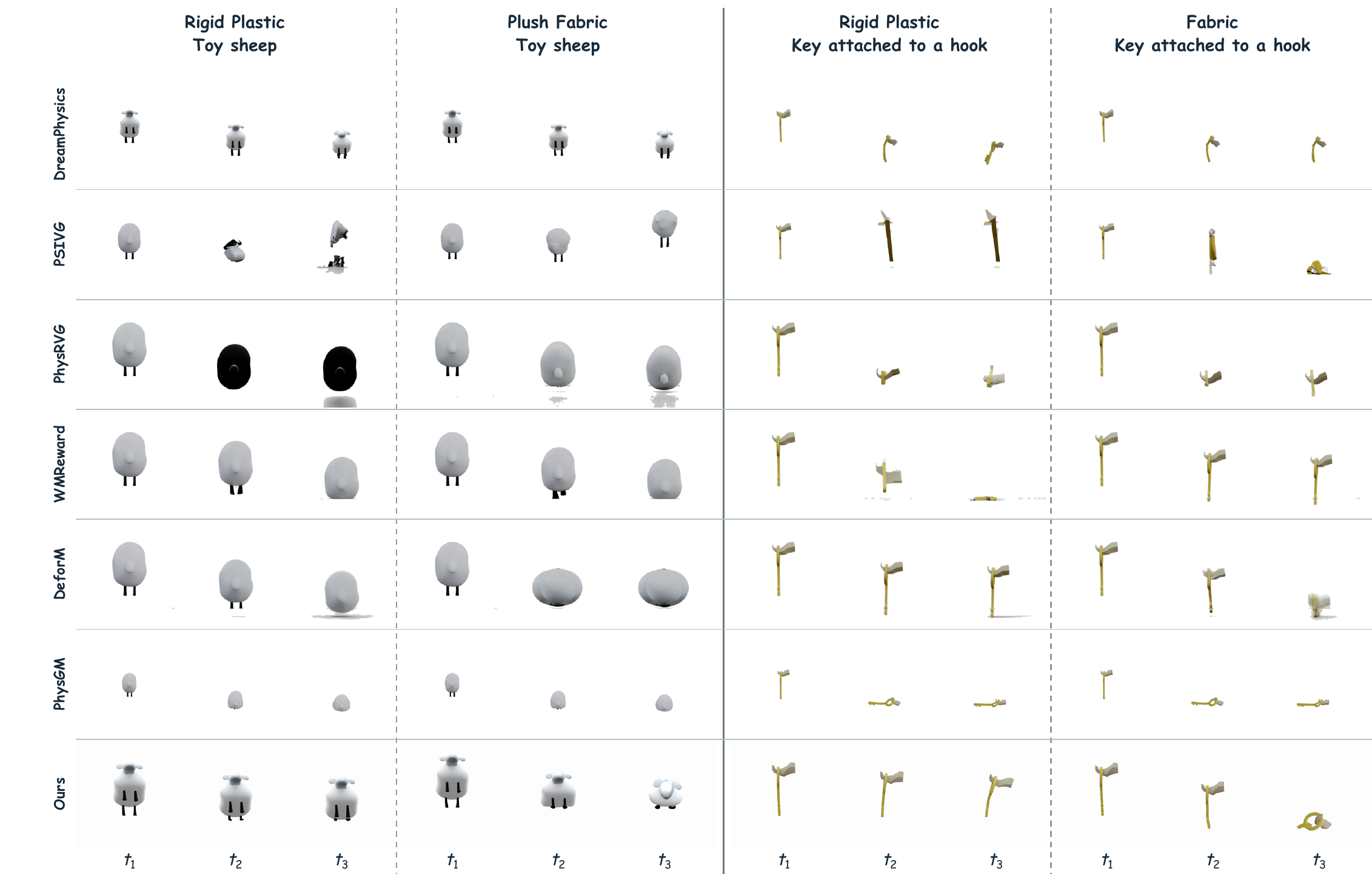}
    \caption{Additional material-controllable comparisons.}
    \label{fig:appendix_controllable_02}
\end{figure}

\begin{figure}[htbp]
    \centering
    \includegraphics[width=\linewidth,keepaspectratio]{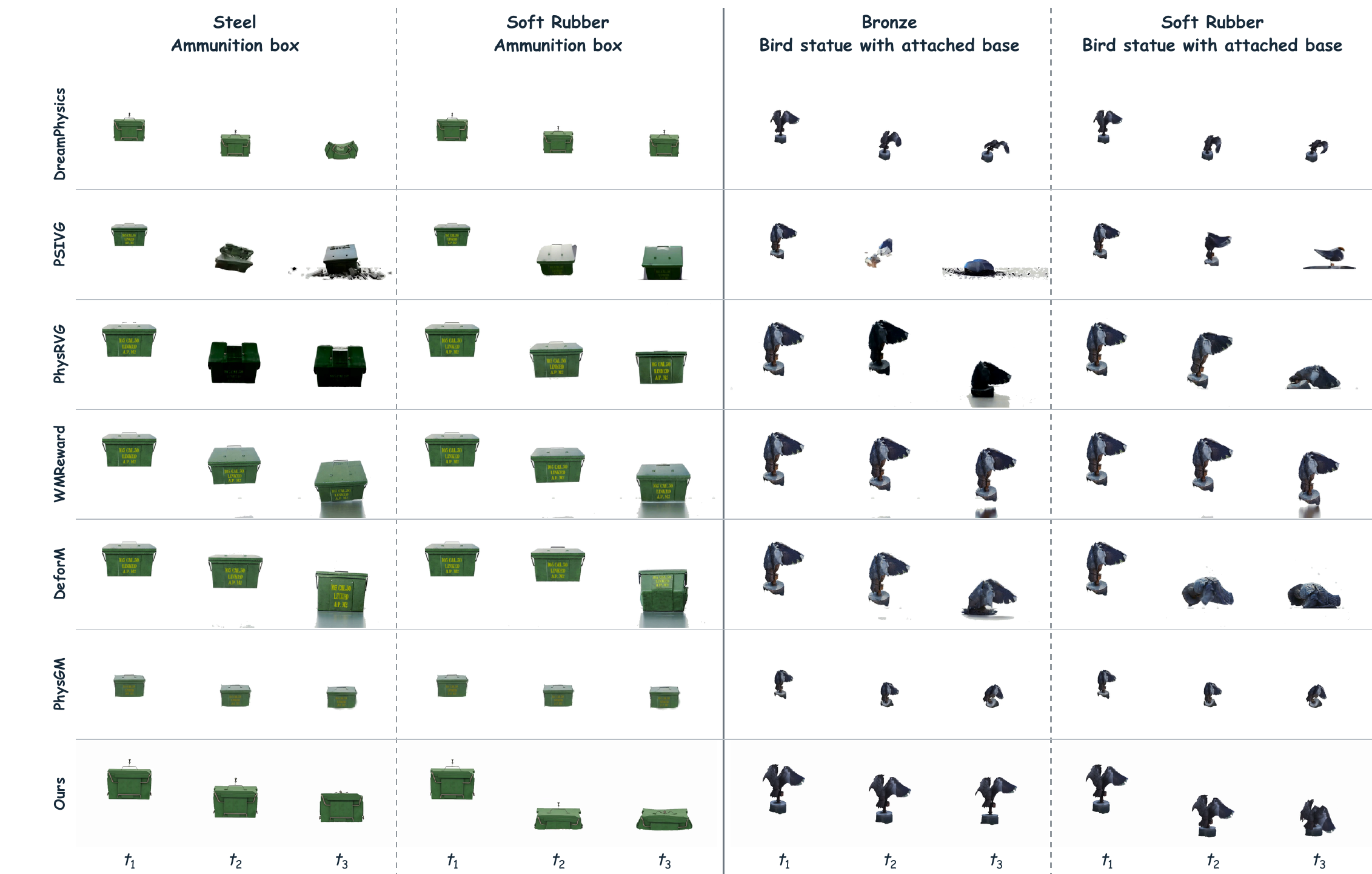}
    \caption{Additional material-controllable comparisons.}
    \label{fig:appendix_controllable_03}
\end{figure}

\begin{figure}[htbp]
    \centering
    \includegraphics[width=\linewidth,keepaspectratio]{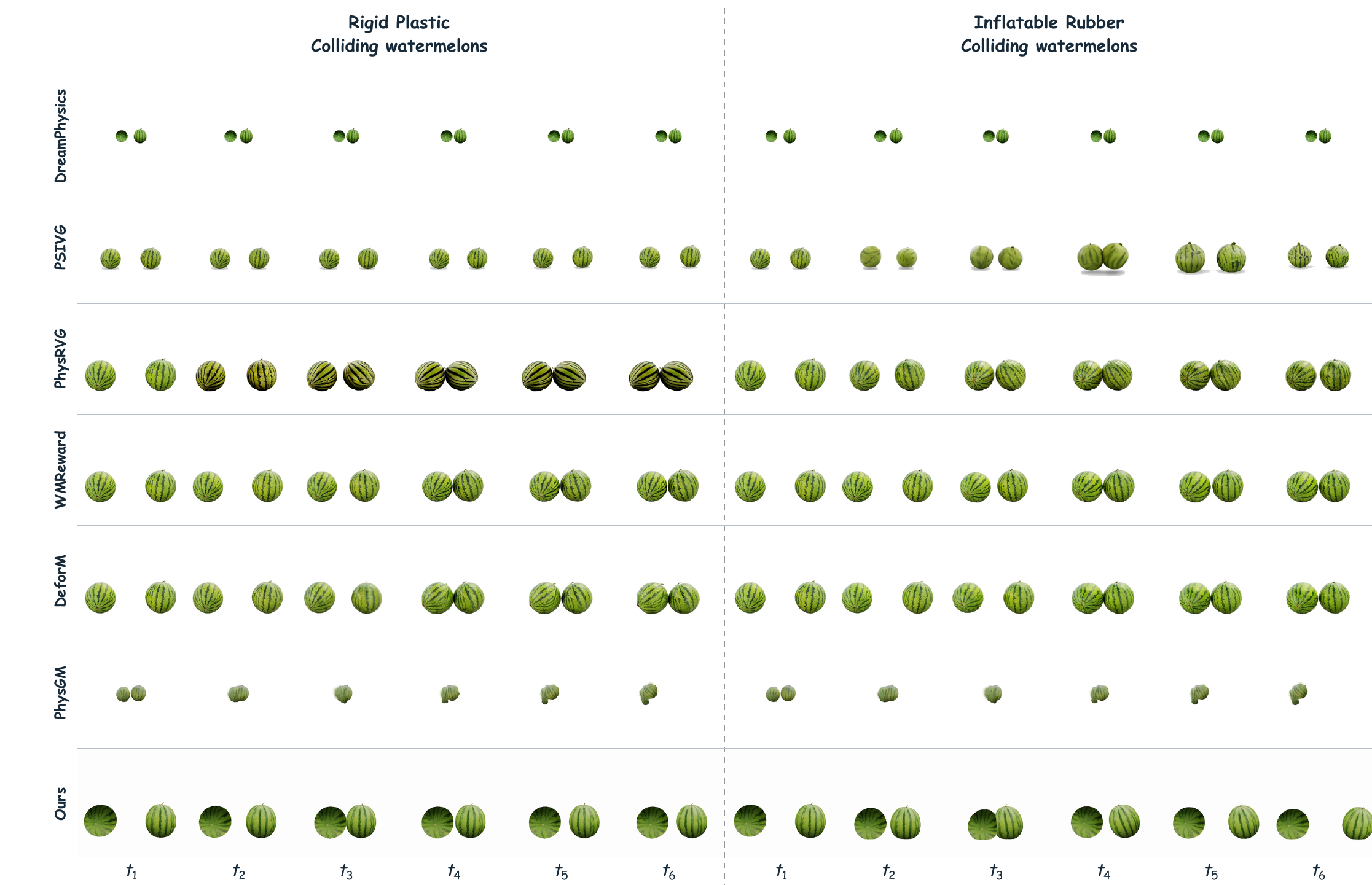}
    \caption{Additional material-controllable comparisons.}
    \label{fig:appendix_controllable_04}
\end{figure}

\begin{figure}[htbp]
    \centering
    \includegraphics[width=\linewidth,keepaspectratio]{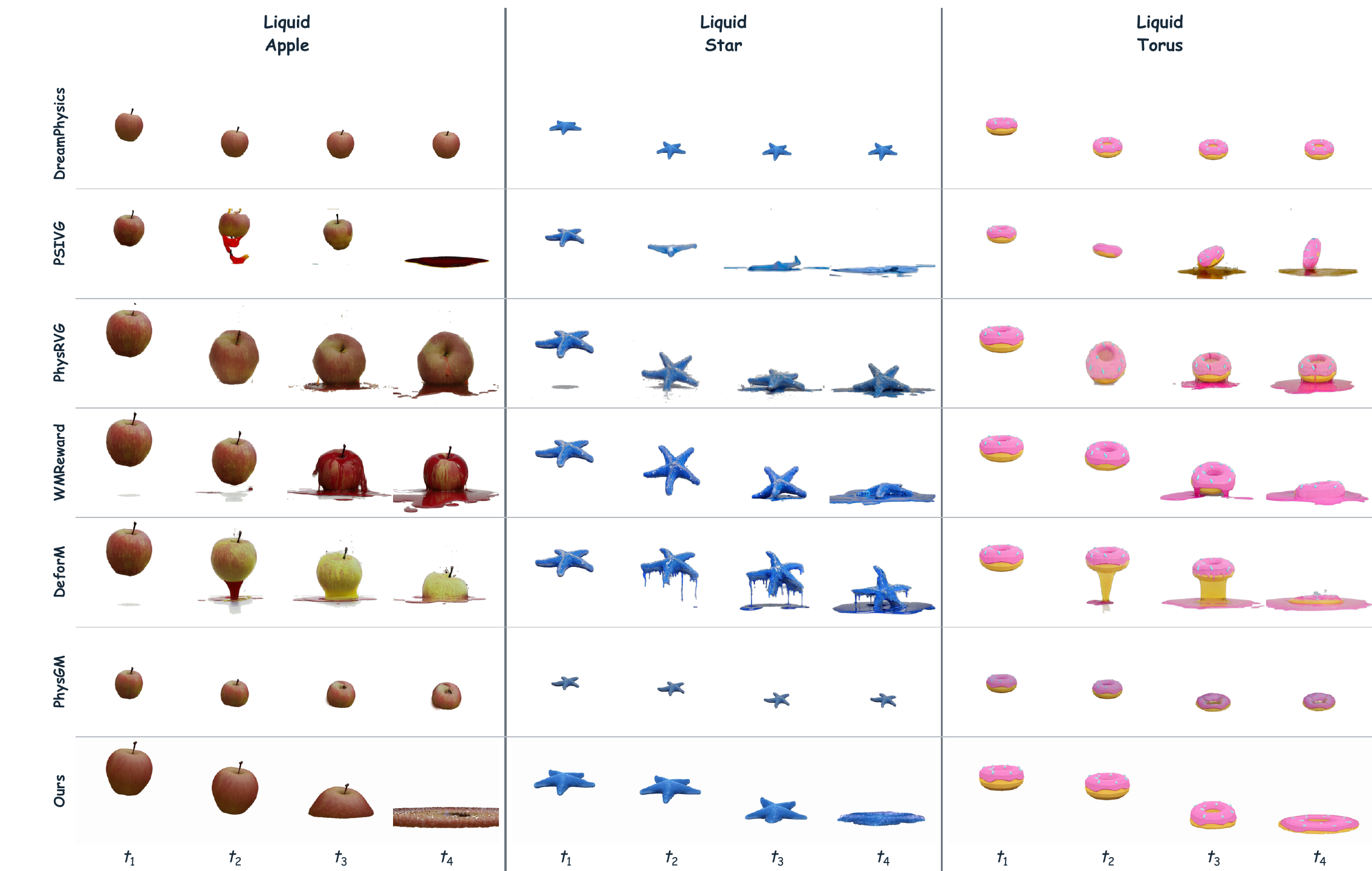}
    \caption{Additional material-controllable comparisons on liquid.}
    \label{fig:appendix_controllable_05}
\end{figure}

\begin{figure}[htbp]
    \centering
    \includegraphics[width=\linewidth,keepaspectratio]{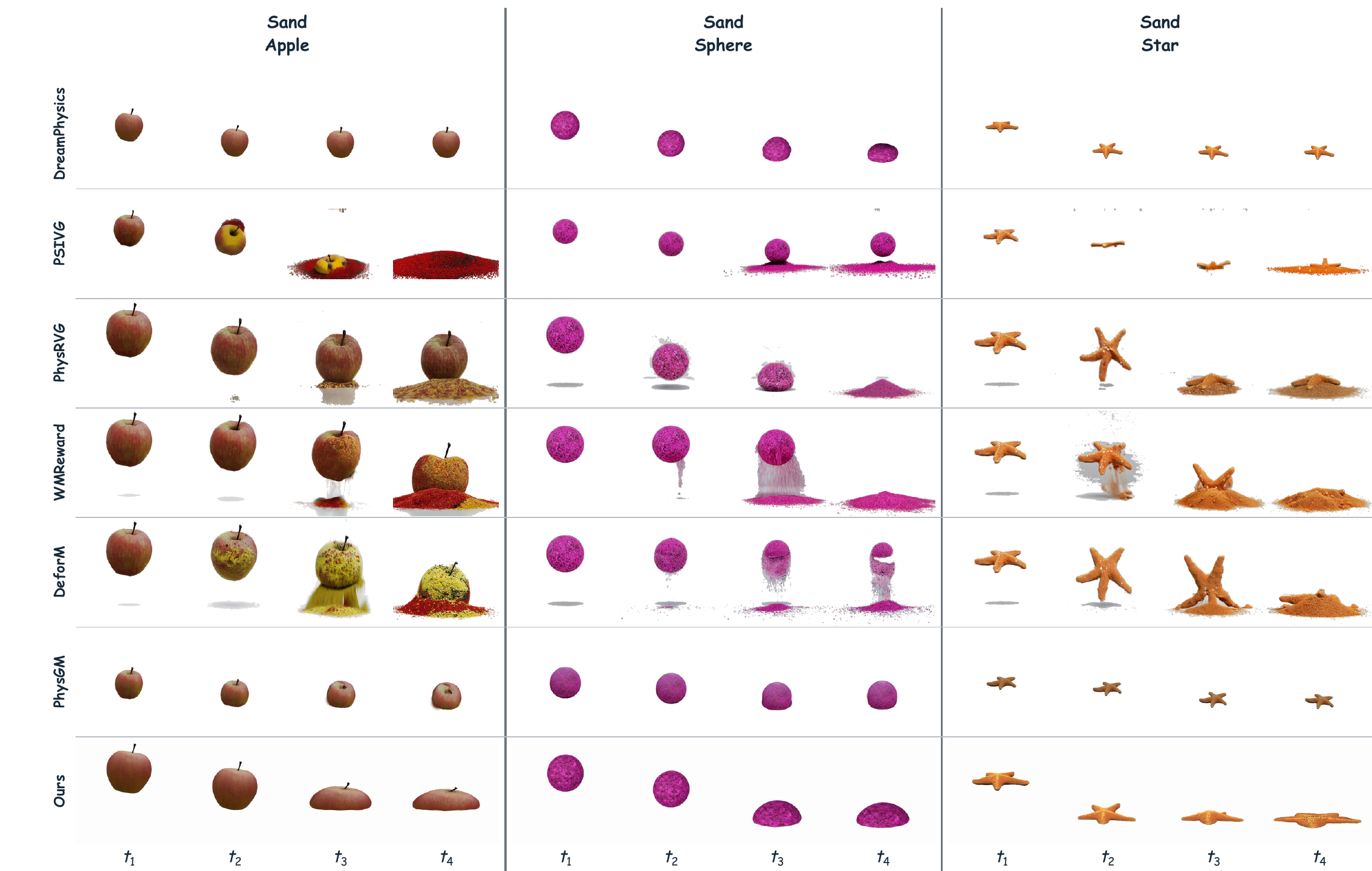}
    \caption{Additional material-controllable comparisons on sand.}
    \label{fig:appendix_controllable_06}
\end{figure}

We present further image results for our method on material-controllable in Figure~\ref{fig:appendix_controllable_01}--\ref{fig:appendix_controllable_06}. Figure~\ref{fig:appendix_controllable_01}--\ref{fig:appendix_controllable_04} demonstrate the results produced by the same object under prompts specifying different materials. Figure~\ref{fig:appendix_controllable_05} and \ref{fig:appendix_controllable_06} shows that our method enables material control for liquid and sand. You can refer to the project link for video and additional examples.

\subsection{More Results on Components Separable}

\begin{figure}[htbp]
    \centering
    \includegraphics[width=\linewidth,keepaspectratio]{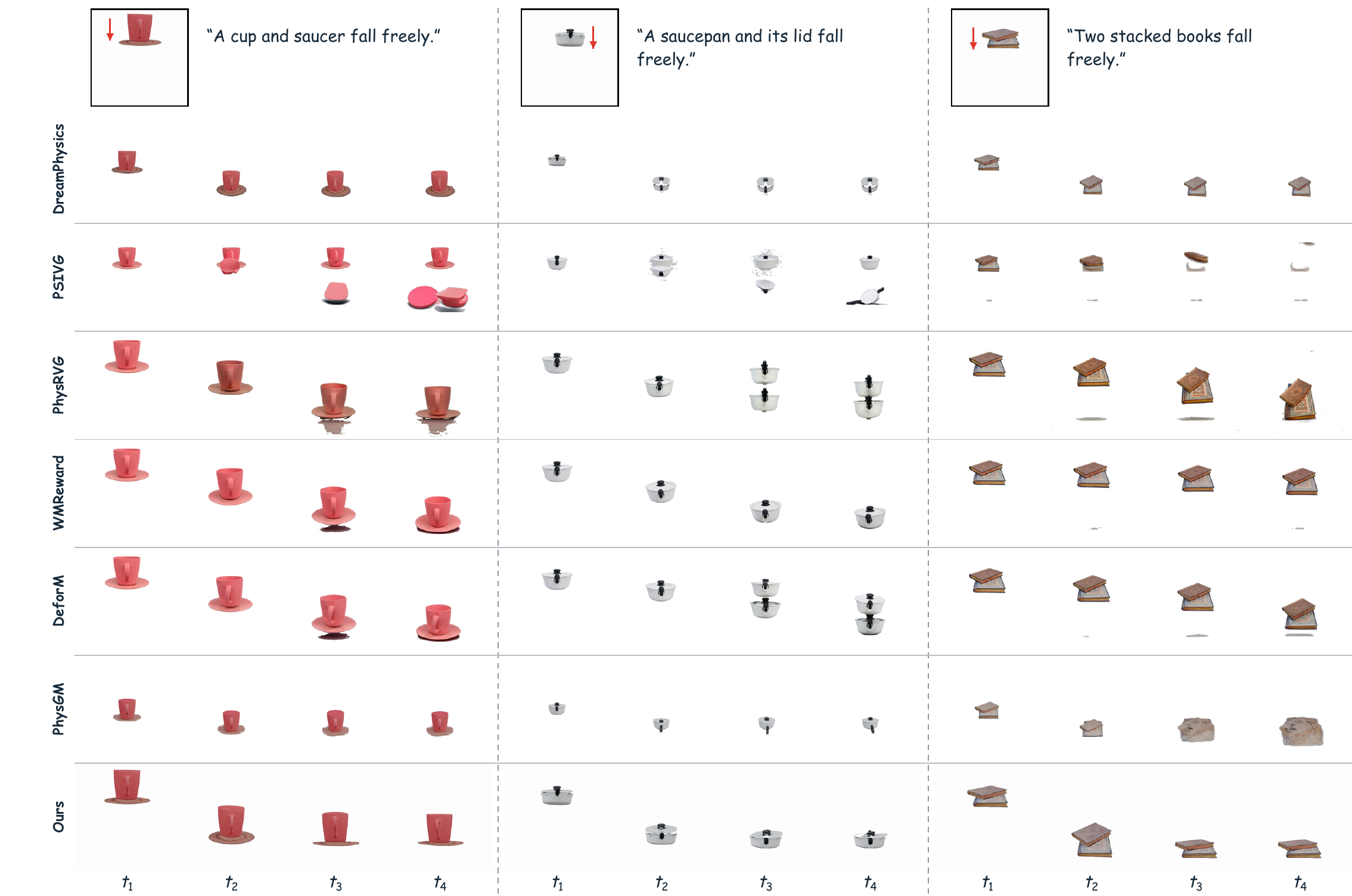}
    \caption{Additional component separable comparisons.}
    \label{fig:appendix_separation_01}
\end{figure}

Figure~\ref{fig:appendix_separation_01} presents further examples of separable components. We can observe relative motion occurring between a cup and its saucer, a pot and its lid, and two books. Please refer to the project link for clearer video and additional examples.

\end{document}